\documentclass[11pt,letterpaper]{article}
\usepackage[T1]{fontenc}
\usepackage{lmodern}

\usepackage{amsmath,amsfonts,bm}

\def\eqref#1{equation~\ref{#1}}

\def\1{\bm{1}}

\DeclareMathAlphabet{\mathsfit}{\encodingdefault}{\sfdefault}{m}{sl}
\SetMathAlphabet{\mathsfit}{bold}{\encodingdefault}{\sfdefault}{bx}{n}

\usepackage[letterpaper,margin=1in]{geometry}
\usepackage{url}
\usepackage{amsmath}
\usepackage{amssymb}
\usepackage{amsthm}
\usepackage{graphicx}
\usepackage{float}
\usepackage{xcolor}
\usepackage{booktabs}
\usepackage{array}
\usepackage{tabularx}
\usepackage[font=small]{caption}
\usepackage{subcaption}
\usepackage{needspace}
\usepackage{natbib}
\usepackage{hyperref}
\hypersetup{hidelinks,pdftitle={Scaling Laws for Generative Reward Models},pdfauthor={Angel Raychev, Nathan Lile, Rafael Rafailov}}

\newif\ifcomments
\commentsfalse

\newcommand{\auditjudge}{\texttt{Qwen3-}\allowbreak\texttt{235B-}\allowbreak\texttt{A22B-}\allowbreak\texttt{Instruct-2507}}
\newcommand{\researchquestion}[1]{%
  \par\smallskip
  \noindent\begingroup
  \setlength{\fboxsep}{5pt}%
  \colorbox{black!4}{%
    \parbox{\dimexpr\linewidth-2\fboxsep\relax}{%
      \raggedright\small\textsc{Research question.}\quad #1%
    }%
  }%
  \endgroup
  \par\smallskip
}

\makeatletter
\newcommand{\subfloat}[2][]{%
  \def\@tempa{#1}%
  \ifx\@tempa\@empty
    #2%
  \else
    \subcaptionbox{#1}{#2}%
  \fi}
\makeatother

\AddToHook{cmd/subsection/before}{\Needspace{9\baselineskip}}
\AddToHook{cmd/section/before}{\Needspace{12\baselineskip}}

\title{Scaling Laws for Generative Reward Models}
\author{
Angel Raychev\\
SynthLabs.ai\\
Stanford University
\and
Nathan Lile\\
SynthLabs.ai
\and
Rafael Rafailov\\
Stanford University\\
Thinking Machines
}
\date{\small Affiliations at the time of the research}

\begin{document}

\maketitle

\begin{abstract}
We study generative reward models (GenRMs) as online preference providers for policy optimization in creative writing. Across Qwen3 reward-model sizes from 0.6B to 14B, Thinking judges trained with GRPO have higher static Gold-label accuracy than Answer-Only judges trained with SFT. This ranking often reverses when the judges train policies through online DPO: Answer-Only rewarders yield higher peak and top-five-average policy Elo in seven of eight matched settings under a fixed, learned 32B Gold evaluator. Intermediate Thinking-judge checkpoints can also outperform later, more statically accurate checkpoints as rewarders. Sweeps over rewarder size, policy size, training duration and DPO regularization show substantial benefits from reward-model capacity, including judges larger than the policy. We analyze trajectories with evaluator-specific Elo arenas, conditional match-resampling intervals and descriptive scaling fits. These results demonstrate an \emph{evaluator--rewarder gap} in the studied pipelines: fixed-distribution judging accuracy does not determine downstream optimization quality. The comparison changes judge-training objective as well as response format, and the main ranking is conditional on its learned reference rather than a universal human-quality result. A limited follow-up using a different learned judge is inconclusive overall.
\end{abstract}

\section{Introduction}

Policies trained from preference feedback can exploit imperfections in their reward models: proxy scores may rise while quality under a separate reference deteriorates \citep{gao2022scaling,rafailov2024rewardoveropt}. Generative reward models (GenRMs) offer a flexible feedback mechanism. By formulating evaluation as conditional text generation rather than scalar prediction, they can leverage pre-training priors and chain-of-thought reasoning and achieve strong performance on static preference benchmarks \citep{mahan2024generative,zhang2024genverifier,whitehouse2025j1}. Whether those stronger evaluators supply better online training signals is a separate question.

Prior work has already shown that static reward-model accuracy can be a poor predictor of downstream optimization \citep{wen2025rethinking,frick2024evaluate,kim2025rethinking}. We study this distinction specifically for generative rewarders as the policy distribution changes, targeting a practitioner-facing question:
\emph{how do judge-training recipe, judge capacity and policy capacity affect the outcomes of online preference optimization?} We study three design dimensions inside the training loop: (i) Thinking GRPO versus Answer-Only SFT judging, (ii) judge scale, and (iii) policy scale.
We instantiate these axes in creative writing using \textsc{LitBench} \citep{fein2025litbench}, a non-verifiable preference domain where correctness cannot be programmatically checked. In such domains, static evaluator accuracy can saturate while reward hacking and proxy-Gold divergence remain salient, making creative writing an ideal testbed for whether stronger static evaluators translate to better in-loop reward signals.

\paragraph{Contributions.}
(1) We present a systematic scaling study of GenRMs deployed as online reward signals, sweeping judge format, judge size (0.6B--14B), policy size (0.6B--8B), training duration and checkpoint selection under online DPO.
(2) We trace policy trajectories under both proxy and Gold evaluators, using within-arena comparisons to separate rising proxy scores from gains under the reference evaluator.
(3) We provide empirical guidance within the studied pipelines: judge scale and training duration matter strongly, Answer-Only rewarders usually sustain higher Gold-rated quality than Thinking variants, and intermediate GenRM checkpoints can outperform later models as rewarders.

Our findings reveal an \emph{evaluator--rewarder gap}. Thinking GenRMs have higher static validation accuracy across the studied sizes, yet this ranking often reverses during online optimization. Answer-Only GenRMs usually yield higher peak Gold Elo, with exceptions where Thinking remains competitive. Additional judge capacity remains valuable above policy size; this does not establish that judging is intrinsically harder than generating.

\begin{table}[htbp]
\centering
\small
\begin{tabularx}{\textwidth}{>{\bfseries\raggedright\arraybackslash}p{0.31\textwidth}>{\raggedright\arraybackslash}X}
\toprule
Observation & Evidence in this paper \\
\midrule
Static evaluation favors Thinking GenRMs & Thinking judges have higher Gold-label validation accuracy across the studied sizes (Figure~\ref{fig:static-eval}). \\
Online rewarder quality is different & In online DPO, Answer-Only judges usually give higher peak and top-five Gold Elo than Thinking judges in the matched sweep (Table~\ref{tab:answer-thinking-summary}, Figure~\ref{fig:answer-thinking-grid}). \\
Judge capacity matters beyond policy size & Larger Answer-Only GenRMs improve Gold-rated policy quality beyond the fixed 4B policy size; Thinking size effects are less monotone (Figure~\ref{fig:genrm-size}). \\
Static accuracy can be over-trained for reward use & Intermediate Thinking checkpoints can outperform fully-trained Thinking checkpoints as online rewarders (Figure~\ref{fig:intermediate-judges}). \\
\bottomrule
\end{tabularx}
\caption{A compact map of the paper's central evidence. The main empirical distinction is between GenRMs as \emph{evaluators} on a fixed distribution and GenRMs as \emph{rewarders} inside an online optimization loop.}
\label{tab:storyline}
\end{table}

\section{Methodology}

We begin with a dataset of human-written stories and implicit preferences inferred from Reddit votes:
\[
\mathcal{D}_{\text{human}} = \{x^{(i)}, y_A^{(i)}, y_B^{(i)}, I_H^{(i)}\}_{i=1}^N,
\]
where $x^{(i)} \in \mathcal{X}$ are prompts, $y_A^{(i)}, y_B^{(i)} \in \mathcal{Y}$ are pairs of stories responding to the same prompt, and $I_H^{(i)} \in \{A, B\}$ denotes the response preferred by the dataset's vote-derived ordering. The subscript ``human'' distinguishes this source supervision from model-generated labels; it does not denote explicit pairwise annotation by human raters.

\textbf{Gold evaluator.} We train a large generative reward model, denoted $V_{\text{gold}}$, on $\mathcal{D}_{\text{human}}$ to obtain a frozen, queryable reference evaluator. Its training target is the source dataset's implicit preference ordering, not an on-demand human judgment. Agreement with these labels on the static story distribution does not establish reliable preferences on later policy-generated responses. We therefore study the downstream behavior of smaller proxy GenRMs \emph{relative to this learned anchor}, not an independently established measure of human writing quality.

To make evaluation computationally feasible, we choose $V_{\text{gold}}$ to be an \emph{Answer-Only} model that outputs a short structured verdict without a reasoning trace. Using $V_{\text{gold}}$, we construct a Gold Preference dataset:
\[
\mathcal{D}_{\text{gold}} = \{x^{(i)}, y_A^{(i)}, y_B^{(i)}, I_G^{(i)}\}_{i=1}^M,
\]
which is then used to train a variety of smaller GenRMs.

\textbf{Proxy GenRMs.} We consider two types of GenRMs: \emph{Answer-Only} and \emph{Thinking}. The Answer-Only models output a direct judgment token ($A$ or $B$):
\[
I \sim v_{\text{ans}}(\cdot \mid x, y_A, y_B),
\]
while the Thinking models first generate a reasoning trace $z$ before producing the final verdict:
\[
z \sim v_{\text{think}}(\cdot \mid x, y_A, y_B),
\quad I \sim v_{\text{think}}(\cdot \mid x, y_A, y_B, z).
\]

\textbf{Training objectives.} We train Answer-Only models using supervised fine-tuning (SFT) and Thinking models using GRPO. SFT directly supervises the verdict; GRPO uses normalized outcome rewards without requiring a reference reasoning trace. This compares two practical training pipelines, rather than isolating the causal effect of reasoning under identical objectives. The 4B Thinking ablation below provides evidence for the chosen GRPO recipe, not an exhaustive comparison of training methods.

The retained GRPO implementation constructs raw outcome rewards in two steps:
\begin{enumerate}
    \item \textbf{Accuracy reward:} A binary reward (1 if the model reaches the correct verdict, 0 otherwise).
    \item \textbf{Raw positional-consistency bonus:} The reward manager links both orderings using a shared pair identifier and computes a majority vote within each ordering's sampled completions. An additional reward of 1 is assigned to correct completions only when both majorities are correct after normalizing response order.
\end{enumerate}
\textbf{Effect of normalization.} In the retained binary-reward setup, the bonus uniformly scales every correct reward within an ordered-prompt group: $r_i=c_g a_i$, with accuracy $a_i\in\{0,1\}$ and $c_g\in\{1,2\}$. GRPO standardizes within that prompt's UID group, so $(r_i-\bar r_g)/(s_g+\epsilon)=(a_i-\bar a_g)/(s_{a,g}+\epsilon/c_g)$. Apart from the numerical stabilizer, the bonus therefore cancels. It changes raw reward statistics but does not provide an independent positional-consistency training signal in this configuration. We preserve the historical implementation and do not attribute the observed GRPO gains to this bonus.
We refer to these trained GenRMs as \emph{Proxy models}, and evaluate them with respect to $V_{\text{gold}}$ preferences.

\textbf{Policy optimization.} We train policies $\pi_\theta$ of varying sizes using \emph{Online Direct Preference Optimization} (Online DPO). GenRMs produce pairwise preferences that DPO can consume directly. Offline DPO reuses a fixed preference dataset; its response-generating policies need not be the frozen reference model in the DPO loss. In online DPO, fresh response pairs are repeatedly sampled from the current policy, so the supervision distribution evolves as the policy changes. This setting lets us study GenRM behavior under policy-induced distribution shift. Prior work has also studied the benefits of online preference optimization \citep{guo2024direct}.

Formally, at online update $t$, we sample $x \sim \mathcal{P}_{\text{train}}$ and two responses $y_1,y_2 \sim \pi_{\theta_t}(\cdot\mid x)$. A proxy GenRM $v$ orders the pair as $(y_v^+,y_v^-)$, where $y_v^+$ is preferred to $y_v^-$. The DPO loss induced by this proxy is
\[
\mathcal{L}_{\text{DPO}}(\theta; v, \pi_{\theta_t})
=
-\mathbb{E}\left[
\log \sigma\left(
\beta
\left[
\log \frac{\pi_\theta(y_v^+\mid x)}{\pi_{\text{ref}}(y_v^+\mid x)}
-
\log \frac{\pi_\theta(y_v^-\mid x)}{\pi_{\text{ref}}(y_v^-\mid x)}
\right]
\right)
\right].
\]
The idealized Gold-aligned update replaces $v$ with $V_{\text{gold}}$. Where the proxy disagrees with Gold matters as well as how often it disagrees: different response pairs contribute different gradient directions and magnitudes. Verdict filtering also changes which sampled pairs reach the update. Static agreement on $\mathcal{D}_{\text{gold}}$ measures neither effect on the evolving pair distribution. Appendix~\ref{app:dpo-noise} gives a short DPO interpretation, without claiming a measured gradient-level explanation of the results.

The policies are trained on prompts sampled from the same broad creative-writing source distribution as the Gold and Proxy datasets, with LitBench prompts excluded by the intended construction procedure. We periodically sample responses on a fixed validation set and estimate Elo ratings from pairwise comparisons under the proxy and Gold evaluators. Each evaluator defines a separate fitted arena: numerical equality between its Elo and another evaluator's Elo is not response-level agreement. Appendix~\ref{app:elo} describes the actual Bradley--Terry fits, checkpoint summaries and conditional uncertainty.

The empirical question is whether a higher-accuracy judge on the fixed Gold-labeled validation set also trains a higher-rated policy under Gold. Agreement on human-story pairs and agreement on evolving policy-generated pairs are different measurements. Our experiments compare the static ranking with downstream policy outcomes; they do not reconstruct trustworthy response-level online agreement from aggregate Elo ratings.

\section{Experimental Setup}

\textbf{Dataset construction.} Our design distinguishes source vote-derived supervision, Gold-derived labels and policy-generation prompts:
\begin{itemize}
    \item \textbf{Source preference dataset $\mathcal{D}_{\text{human}}$:} \textsc{LitBench} \citep{fein2025litbench} curates same-prompt Reddit story pairs using vote margins, engagement, posting-time and length filters. The higher-voted story defines an implicit preference label, rather than an explicit human pairwise annotation. The retained Gold-training script uses rows 1{,}000--21{,}999 for training (21{,}000 pairs) and rows 0--999 for validation (1{,}000 pairs).
    \item \textbf{Gold preference dataset $\mathcal{D}_{\text{gold}}$:} The frozen Gold evaluator relabels pairs of human-written stories for proxy training. The retained artifact contains 21{,}827 raw training rows and 1{,}000 validation rows. Its training count is consistent with the remaining \textsc{LitBench} subset after row 21{,}999, although the complete original training-relabel invocation is not retained. The validation pairs reuse the first 1{,}000 source pairs, with Gold-derived rather than vote-derived labels. Thus Gold and proxy validation share story pairs; they are not independent source samples. Proxy training filters null Gold labels and expands surviving pairs into both positional orders.
    \item \textbf{Policy prompts and validation references:} Policy training uses prepared \textsc{WritingPrompts} prompts\footnote{\url{https://huggingface.co/datasets/euclaise/WritingPrompts_preferences}}, with \textsc{LitBench} prompts excluded by the construction procedure. The retained main preparation combines 50{,}000 train, 20{,}000 reserve and 20{,}000 extra prompts; a separate supplementary split supports later continuations. Policy validation instead uses the prompts from the first 1{,}000 \textsc{LitBench} pair records, with each original higher-voted story as its fixed response reference. This reuses the static validation records, not Gold-training examples; it is not a fresh \textsc{WritingPrompts} sample. Online training preferences come from newly generated policy response pairs judged by the proxy, not from the Reddit stories.
\end{itemize}

This separation identifies three distinct roles: source labels train Gold, Gold-derived labels train proxy judges, and fresh proxy-labeled policy responses drive online DPO. The frozen reference policy in the DPO loss is a fourth role, distinct from the fixed Reddit story used in validation.

\textbf{Model families.} We use the \textsc{Qwen3} model family for all GenRMs and policies. It provides Thinking and Answer-Only modes across several parameter sizes, allowing matched mode/size comparisons within one family. The comparison still changes judge-training objective, and parameter scale is not isolated from all model-specific differences. Using the same family for policies and GenRMs lets us study their capacity interaction under a consistent experimental framework.

\textbf{Gold GenRM training.} We train the Gold evaluator from \textsc{Qwen3-32B} using supervised fine-tuning on $\mathcal{D}_{\text{human}}$. To improve positional robustness, we train on both orderings $(y_A, y_B)$ and $(y_B, y_A)$, resulting in 42{,}000 ordered training examples before any input filtering. We use a batch size of 128 and a learning rate of $1 \times 10^{-5}$.

\textbf{Proxy GenRM training.} We train Proxy GenRMs from the \textsc{Qwen3} series at sizes \{0.6B, 1.7B, 4B, 8B, 14B\} using $\mathcal{D}_{\text{gold}}$. We refer to off-the-shelf models as \texttt{Baseline-[size]-Ans} and \texttt{Baseline-[size]-Think}. Answer-Only models use supervised fine-tuning with batch size 128 and learning rate $1 \times 10^{-5}$. Each retained Gold-labeled pair produces two ordered examples; the final consumed count depends on label and input filtering. Thinking models use GRPO with 16 prompts per step, 8 completions per prompt, configured minibatch size 64 and learning rate $1 \times 10^{-6}$. The retained launcher disables the trainer-level KL reward controller but enables a separate actor KL loss with coefficient $3\times10^{-5}$; this is not a zero-KL setup. We refer to trained proxies as \texttt{GenRM-[size]-Ans} and \texttt{GenRM-[size]-Think}.

\textbf{Policy training.} We train policies $\pi_\theta$ from the \textsc{Qwen3} series at sizes \{0.6B, 1.7B, 4B, 8B\} using Online DPO with thinking disabled. We use a prompt batch size of 192, global minibatch size of 64 preference pairs, learning rate $1 \times 10^{-6}$ and $\beta = 0.02$ unless varied. Two current-policy responses are generated per prompt. The proxy judges both orderings, and the trainer sums order-normalized verdicts, discarding zero-sum pairs. A verdict parsed as invalid contributes zero, so one valid verdict and one invalid verdict can still yield a retained pair; this is not a guarantee of two valid agreeing votes. Policy and reference likelihoods use the same historical response mask, which excludes the final attended token, including a content token if a capped generation lacks EOS. The original experiment description specifies prompt and response caps of 512 and 2048 tokens, respectively. The retained hand-edited launcher instead contains caps of 500 and 1500; resolved per-run configurations have not been recovered for every plotted trajectory, so those original caps should be read as nominal settings rather than a newly verified run-wide invariant.

\textbf{Evaluation.} During policy training, we save validation generations every 10 training steps; the retained launcher saves model weights every 50 steps. We call the evaluated policy states evaluation snapshots, without implying that every state has separately retained weights. We compute Elo ratings from pairwise comparisons judged by the proxy and Gold evaluators. Each evaluator has its own fitted ratings over the relevant contestants and comparisons; axes from different evaluators are not one shared numerical scale. The principal rating exports cover approximately 460 steps, usually 47 evaluation snapshots including step zero. Subsequent policy continuations extend beyond this window; they do not share a newly reconstructed common-arena rating table and are not spliced into these curves.

\textbf{Artifact reconstruction.} Counts above distinguish stored raw splits from examples retained after filtering. The inspected code identifies the consumed data roles, but does not retain every original preparation invocation or resolved per-run configuration. The released analysis identifies its numerical inputs separately from the historical training recipes.

\textbf{Code and analysis.} Training implementation, historical recipes, numerical results and figure-reproduction instructions are available in the \href{https://github.com/RaychevAngel/generative-reward-models}{public research repository}. The release distinguishes reproducible analysis of retained results from full training replay, which additionally requires model and dataset access. Prompt templates are provided in Appendix~\ref{sec:prompts}.

\begin{table}[t]
\centering
\small
\begin{tabularx}{\textwidth}{>{\bfseries}p{0.22\textwidth}X>{\raggedright\arraybackslash}p{0.18\textwidth}}
\toprule
Axis & Values swept & Default / anchor \\
\midrule
Proxy GenRM format & Answer-Only SFT, Thinking GRPO & Both compared \\
Proxy GenRM size & 0.6B, 1.7B, 4B, 8B, 14B & 4B unless varied \\
Policy size & 0.6B, 1.7B, 4B, 8B & 4B unless varied \\
Online DPO $\beta$ & 0.1, 0.02, 0.01, 0.005, 0.002 & 0.02 \\
Training trajectory & Policy evaluation every 10 steps; intermediate Thinking GenRM checkpoints & Principal exports: about 460 steps \\
\bottomrule
\end{tabularx}
\caption{Axes studied in matched and fixed-axis online-DPO sweeps, not a full factorial grid. The design separates fixed-distribution accuracy from in-loop rewarder behavior.}
\label{tab:experiment-grid}
\end{table}

\section{Experiments and Results}\label{sec:experiments}

\subsection{GenRM Training}
\begin{figure}[ht]
    \centering
    \includegraphics[width=0.85\linewidth]{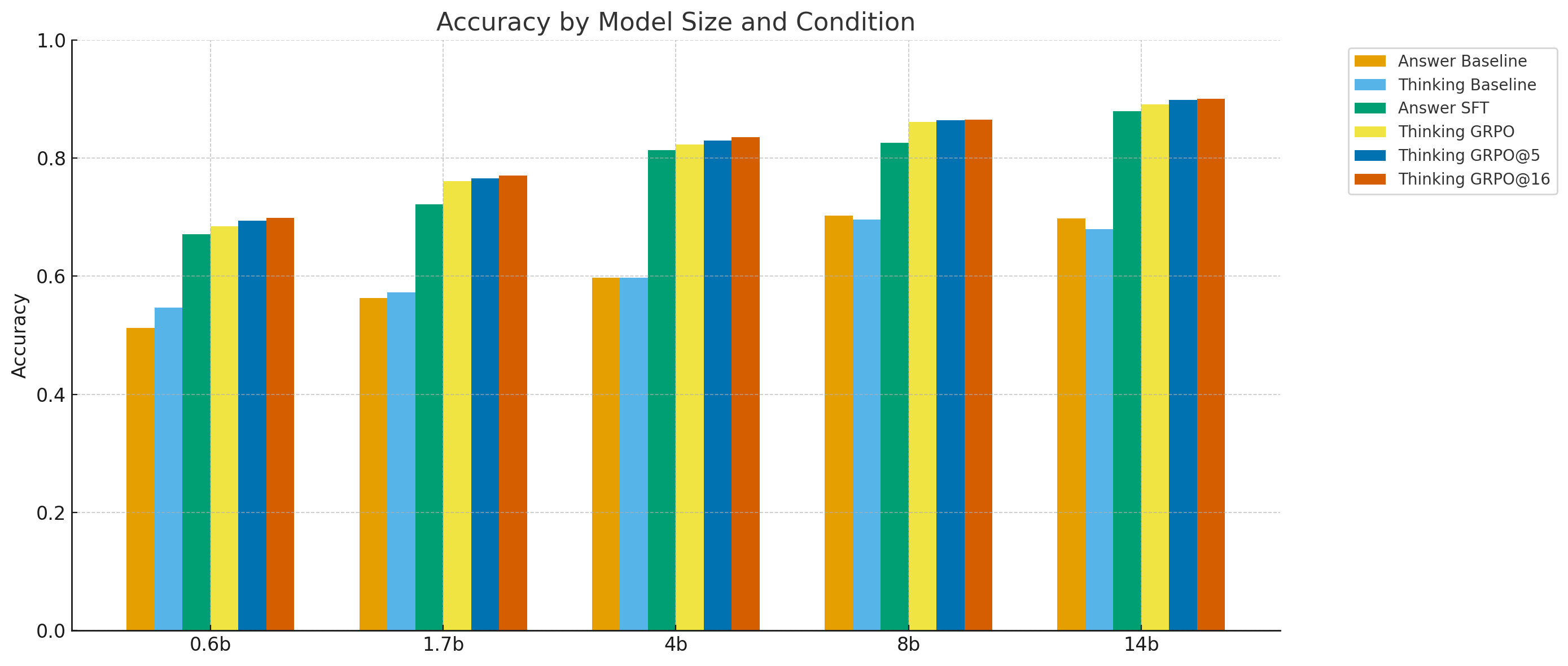}
    \caption{Static evaluator accuracy on the Gold Preference dataset. Thinking GenRMs exceed Answer-Only models across the studied sizes, while both trained variants substantially exceed off-the-shelf baselines. The size of the static advantage varies by model.}
    \label{fig:static-eval}
\end{figure}

\textbf{Gold evaluator validation.} The Gold GenRM achieves 79\% agreement with \textsc{LitBench}'s vote-derived labels on the held-out validation split of $\mathcal{D}_{\text{human}}$. This establishes its agreement with the dataset ordering on source story pairs, not human agreement on policy-generated outputs after optimization. The source study also reports strong creative-writing preference prediction \citep{fein2025litbench}, but its different evaluators and procedures are not a matched baseline for this number.

\textbf{Trained GenRMs substantially outperform baselines.} Both Answer-Only SFT and Thinking GRPO improve static Gold-label accuracy over their off-the-shelf counterparts (Figure~\ref{fig:static-eval}). Similar baseline accuracies do not rule out evaluator-specific preferences or format bias. The trained-versus-baseline improvement is substantial across the sizes studied.

\textbf{GRPO is effective in the 4B Thinking ablation.} On the Gold Preference dataset, the off-the-shelf model achieves 60.0\%, STaR achieves 60.7\%, distillation from \textsc{Qwen3-32B} thinking traces (SFT-style) achieves 64.7\%, and the retained GRPO recipe achieves 83.2\%. GRPO gives the largest improvement among these tested recipes, motivating its use for the Thinking proxies. The raw order-consistency bonus does not supply a distinct normalized update, as explained above; this is not evidence for its separate benefit.

\textbf{Thinking GenRMs outperform Answer-Only models on in-distribution evaluation.} Thinking models have higher static Gold-label accuracy at each studied size (Figure~\ref{fig:static-eval}); the gap varies by size. Majority voting over $k=5$ and $k=16$ Thinking completions gives modest further gains. Given the additional computation, we use one completion per judged ordering during policy training.

\begin{figure}[ht]
    \centering
    \includegraphics[width=0.75\textwidth]{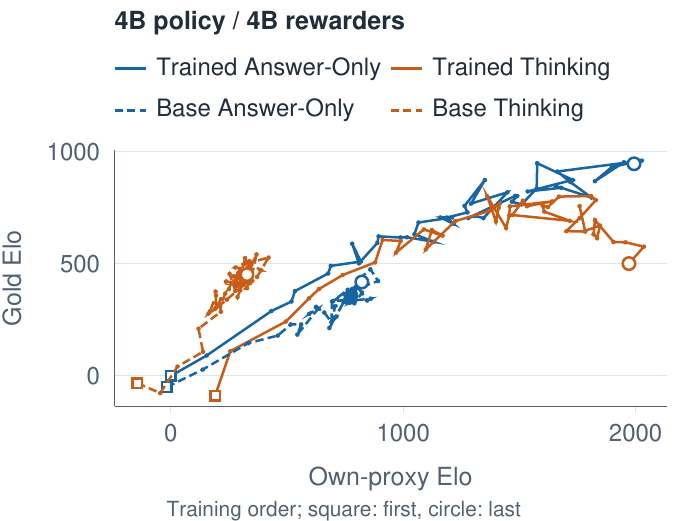}
    \caption{Trained versus off-the-shelf 4B rewarders training a 4B policy. Points are recorded snapshots connected in policy-training order; squares/circles mark first/last observations, not maxima. Gold is shared within this experiment family, whereas own-proxy ratings are separate evaluator fits. The chronological Gold comparison in Appendix~\ref{app:empirical} exposes the late trained/baseline Thinking crossover.}
    \label{fig:rewarder-gap}
\end{figure}

\begin{figure}[htbp]
    \centering
    \includegraphics[width=\textwidth]{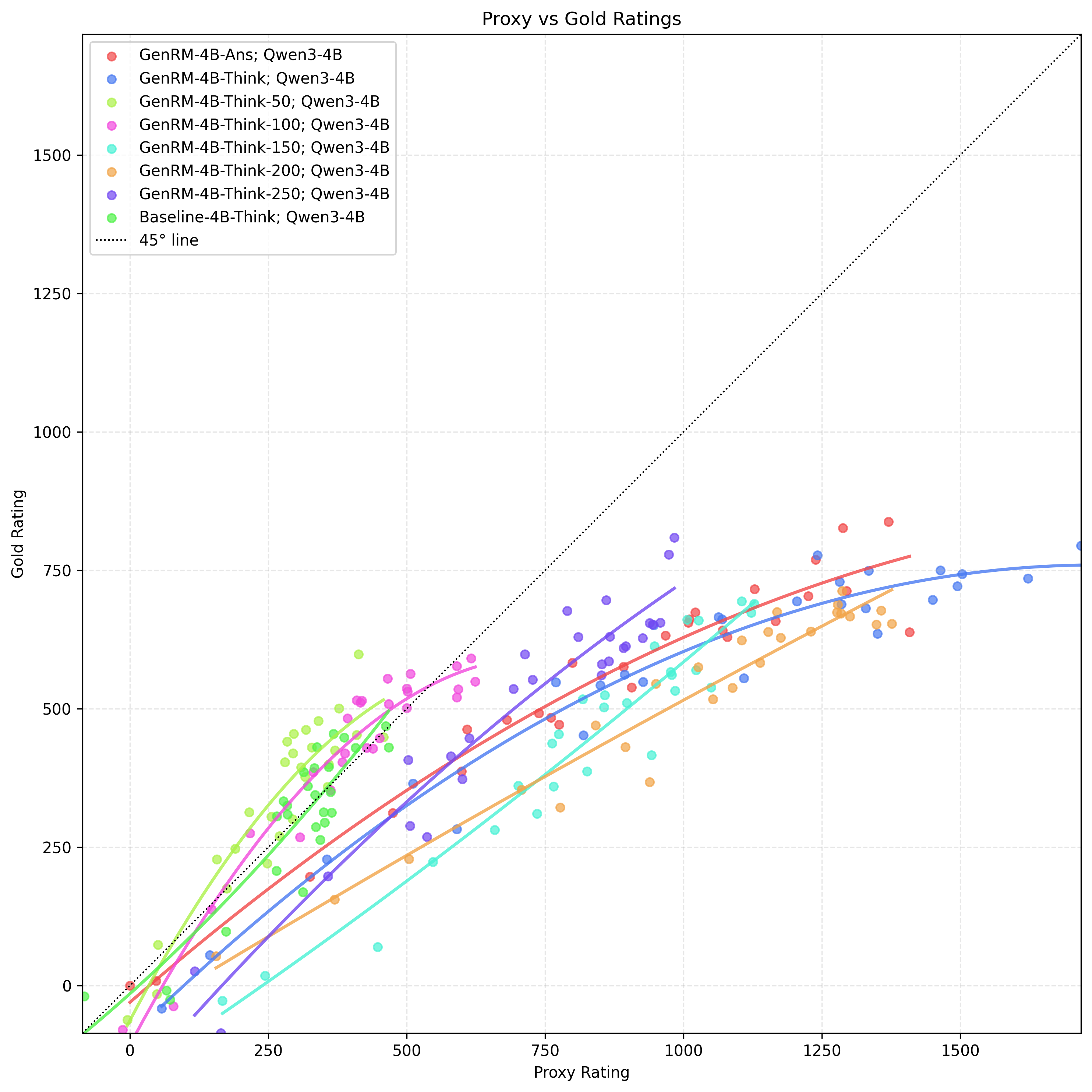}
    \caption{Policies trained with intermediate 4B Thinking-judge checkpoints. The retained historical image uses fitted display curves, not chronological connecting lines; its diagonal is not verdict agreement. The original account specifies 40{,}000 policy-training prompts and the judge-checkpoint accuracies reported in the text. This separate ablation illustrates that the most accurate static evaluator need not be the strongest online rewarder.}
    \label{fig:intermediate-judges}
\end{figure}

\subsection{Trained GenRMs vs.\ Baseline Models in Policy Training}
\researchquestion{Does training the GenRM improve downstream policy quality compared to off-the-shelf models?}
We compare policies optimized with trained versus off-the-shelf 4B GenRMs, with the policy also fixed at 4B (Figure~\ref{fig:rewarder-gap}). Here ``baseline'' describes the rewarder, not an untrained policy. Trained rewarders produce higher peak Gold ratings in both modes. That is not uniform dominance at every training duration: at matched step 460, the off-the-shelf Thinking rewarder's policy reaches 542 Gold Elo versus 501 for the trained Thinking rewarder's policy, after the latter declines from a higher peak of 803. Answer-Only retains a trained-rewarder advantage throughout the matched trajectory, with peaks 961 versus 476. The restored chronological comparison in Appendix~\ref{app:empirical} makes the late crossover and the peak/final distinction explicit. While Thinking has the stronger off-the-shelf rewarder peak, Answer-Only has the stronger trained-rewarder peak in this arena.

\subsection{Answer-Only vs Thinking}
\researchquestion{Which of the compared Answer-Only SFT and Thinking GRPO pipelines provides the better online rewarder?}
\begin{figure}[htbp]
    \centering
    \includegraphics[width=\textwidth]{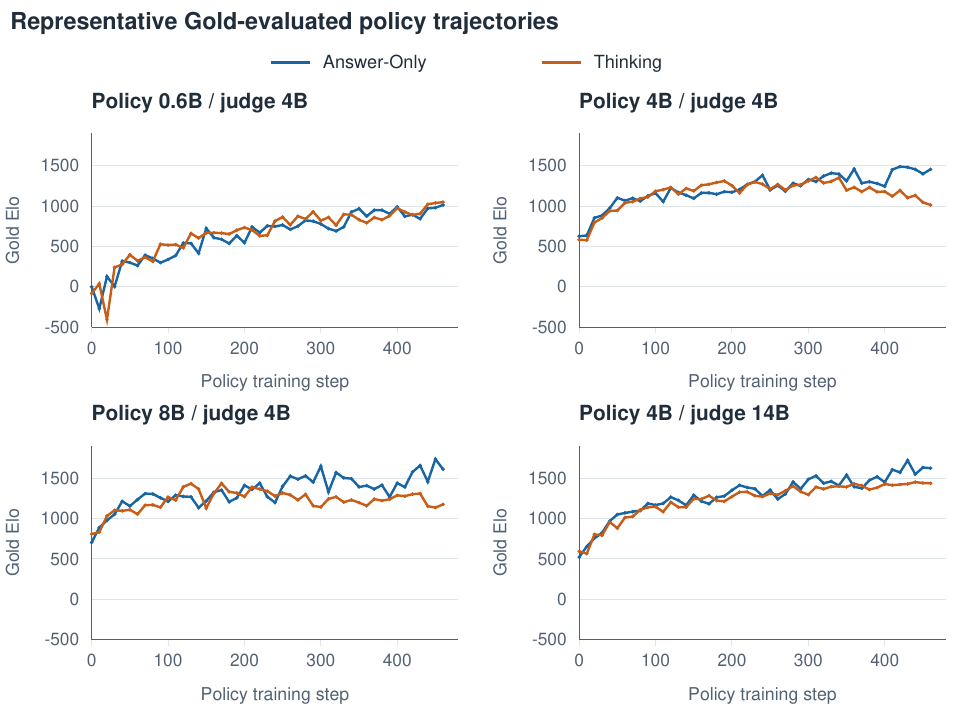}
    \caption{Four representative Gold-rated policy trajectories for Answer-Only SFT (blue) and Thinking GRPO rewarders (orange), including the smallest-policy exception, the 4B anchor and larger-policy/rewarder cases. All panels share the global Gold arena and vertical limits; lines connect recorded training snapshots, not fitted curves. All eight matched configurations are summarized in Table~\ref{tab:answer-thinking-summary} and shown chronologically in Appendix~\ref{app:empirical}.}
    \label{fig:answer-thinking-grid}
\end{figure}

\begin{table}[t]
\centering
\small
\begin{tabular}{llll}
\toprule
Policy & GenRM & Peak gap [95\% CI] & Top-5 gap [95\% CI] \\
\midrule
0.6B & 4B & $-35$ [$-134$, $49$] & $-19$ [$-59$, $25$] \\
1.7B & 4B & $+113$ [$43$, $176$] & $+94$ [$48$, $134$] \\
4B & 0.6B & $+41$ [$-32$, $104$] & $+24$ [$-2$, $65$] \\
4B & 1.7B & $+145$ [$104$, $239$] & $+159$ [$133$, $214$] \\
4B & 4B & $+135$ [$49$, $241$] & $+141$ [$93$, $189$] \\
4B & 8B & $+41$ [$-67$, $143$] & $+25$ [$-18$, $88$] \\
4B & 14B & $+269$ [$128$, $488$] & $+193$ [$134$, $257$] \\
8B & 4B & $+302$ [$158$, $573$] & $+243$ [$189$, $343$] \\
\bottomrule
\end{tabular}
\caption{Matched Answer-Only versus Thinking Gold Elo gaps in the global all-configuration arena, reported as Answer-Only minus Thinking. Point gaps and intervals use the lightly regularized refit and 200 bootstrap resamples described in Appendix~\ref{app:elo}; plotted trajectories use the original exported ratings. Answer-Only wins 7 of 8 matched cells under both summaries; the exception is the smallest-policy setting with a 4B judge.}
\label{tab:answer-thinking-summary}
\end{table}

In the shared global Gold arena, Answer-Only rewarders give higher peak policy Elo in 7 of 8 matched configurations (Table~\ref{tab:answer-thinking-summary}). The exception is the 0.6B policy with a 4B judge: Thinking peaks at 1048 versus 1013 for Answer-Only. Averaging each run's five highest-rated snapshots gives the same 7-of-8 pattern, with mean gaps of 107 Elo for top-five averages and 126 for peaks. Under 200 independent-row bootstrap resamples and regularized refits, 5 of 8 peak gaps and 5 of 8 top-five gaps have 95\% percentile intervals above zero. This is conditional match-table uncertainty, not prompt-cluster, generation or retraining uncertainty (Appendix~\ref{app:elo}). Figure~\ref{fig:answer-thinking-grid} shows representative Gold-rated training trajectories; Appendix~\ref{app:empirical} provides the complete matched set, including competitive Thinking settings.
The reversal matters because static evaluation favors Thinking, and the off-the-shelf baseline comparison also differs from the trained comparison. Under the learned Gold reference, the statically stronger practical judge-training pipeline is not generally the better online reward provider.
The difference is consistent with different responses to policy-induced distribution shift, but does not identify that mechanism: Gold's own preferences can also change with its inputs, and both judge training and response mode differ. Two observations help locate the gap without resolving its cause.

First, higher in-distribution GenRM accuracy does not necessarily translate to better policy training. Figure~\ref{fig:intermediate-judges} compares intermediate 4B Thinking checkpoints at 0, 50, 100, 150, 200, 250 and 300 GRPO steps. The contemporaneous experiment account reports static accuracies of 59.7\%, 61.5\%, 64.5\%, 68.7\%, 71.5\%, 79.2\% and 82.3\%, respectively. These are the checkpoint-ablation measurements, not the separate 83.2\% recipe comparison above. The 250-step judge (79.2\%) produces a stronger online rewarder than the later, more statically accurate checkpoint in this ablation. This illustrates that evaluator-optimal and rewarder-optimal checkpoint selection can differ. This late ablation used 40{,}000 policy-training prompts rather than the main 90{,}000.

Second, GRPO supervises final preferences rather than the correctness of each reasoning step. A Thinking GenRM may therefore learn traces that support training-distribution labels without remaining reliable under policy-induced shift; we treat this as a hypothesis, not a measured mechanism. The trajectory data show differing stability: across the eight matched global configurations, Answer-Only runs lose an average of 60 Elo from peak to final, compared with 130 for Thinking. An auxiliary verdict-agreement reconstruction could align rows only by checkpoint-player pairs and row order, not unique prompt/response identifiers. We do not use its percentages as verified response-level agreement or as mechanistic evidence.

These results should not be read as evidence that Answer-Only supervision preserves stylistic diversity. A surface audit of stored peak-checkpoint generations in two 4B-policy cells shows that high Gold Elo can coexist with template convergence: in the 4B-policy/4B-judge cell, 94\% of Answer-Only peak outputs share the same opening line, and the duplicate-line rate is higher than in the corresponding Thinking peak outputs. The stronger Answer-Only result is therefore a preference-optimization result, not a diversity result.
In summary, Answer-Only rewarders require shorter verdict generations and, in these runs, usually sustain higher Gold-rated policy quality. This does not independently establish greater human-quality robustness or identify reasoning as the cause. Intermediate Thinking checkpoints remain useful candidates for rewarder selection.

\subsection{Effect of GenRM Size}
\researchquestion{Does scaling the reward model improve policy quality, and do gains continue when the GenRM exceeds the policy in size?}
\begin{figure}[t]
    \centering
    \subfloat[Answer-Only GenRMs]{\includegraphics[width=0.47\textwidth]{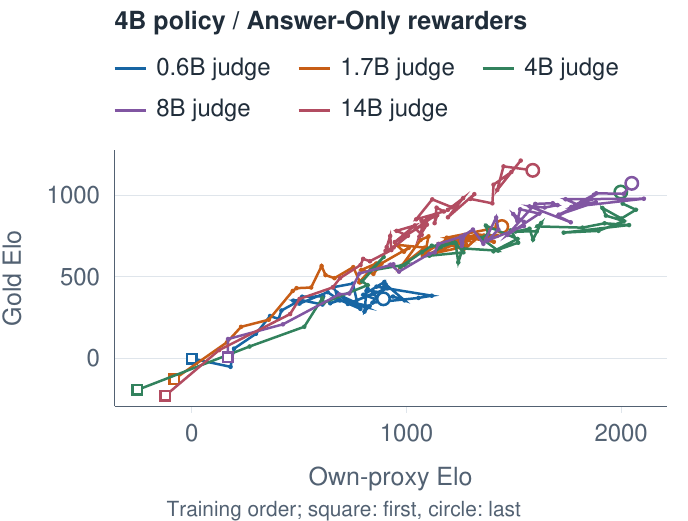}}
    \hfill
    \subfloat[Thinking GenRMs]{\includegraphics[width=0.47\textwidth]{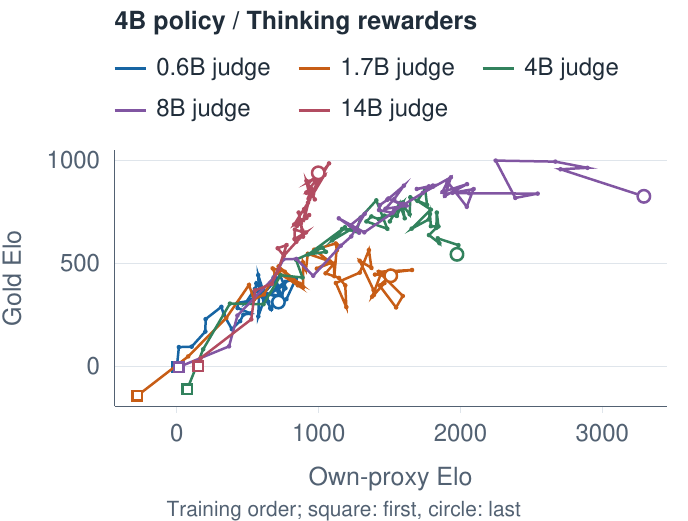}}
    \caption{Rewarder-size sweeps with a fixed 4B policy, using the retained family-specific Gold arenas. Points are recorded snapshots connected in training order; squares/circles mark first/last observations. Own-proxy ratings come from separate rewarders and are not one calibrated scale. Larger Answer-Only rewarders improve Gold Elo beyond policy size; Thinking scaling is less monotone. The common-global-arena comparison is in Appendix~\ref{app:empirical}.}
    \label{fig:genrm-size}
\end{figure}
We investigate the effect of scaling GenRM size while fixing the policy size at 4B.
Figure~\ref{fig:genrm-size} shows results for both Answer-Only and Thinking GenRMs across scales from 0.6B to 14B.
We observe clear gains from increasing GenRM size, especially for Answer-Only rewarders. With a 4B policy, the Answer-Only 14B judge is the strongest judge-size configuration in the global arena. Thinking rewarders also benefit substantially from scale, but less monotonically: the 8B Thinking judge slightly exceeds the 14B Thinking judge in peak Gold Elo (1507 vs.\ 1451) and top-5 average Gold Elo (1488 vs.\ 1437). Thus additional rewarder capacity can remain useful above policy size, while its benefit depends on the judge's training and online behavior. Parameter counts alone do not establish a general ranking of the difficulty of judging and generating.

Appendix~\ref{app:scaling-law-fits} summarizes the existing sweep with descriptive scaling fits. A log-linear model using policy size, judge size and judge mode has an in-sample $R^2$ of 95.2\% for top-5 Gold Elo across the 16 matched global configurations. In that fit, the coefficients correspond to 176 Elo per Answer-Only policy-size doubling and 136 per judge-size doubling; the Thinking-run values are 119 and 116 Elo. These are fitted summaries of this sparse grid, not predicted causal gains from a new doubling or a compute-optimal allocation rule.

\begin{table}[t]
\centering
\small
\begin{tabularx}{\textwidth}{Xrrr}
\toprule
Configuration & Top-5 avg Gold Elo & Peak Gold Elo & Peak step \\
\midrule
GenRM-4B-Ans; Policy 8B & 1645 & 1736 & 450 \\
GenRM-14B-Ans; Policy 4B & 1630 & 1719 & 430 \\
GenRM-8B-Ans; Policy 4B & 1513 & 1549 & 460 \\
GenRM-8B-Think; Policy 4B & 1488 & 1507 & 460 \\
GenRM-4B-Ans; Policy 4B & 1467 & 1488 & 420 \\
GenRM-14B-Think; Policy 4B & 1437 & 1451 & 440 \\
\bottomrule
\end{tabularx}
\caption{Top online-DPO configurations in the global all-configuration arena, ranked by average Gold Elo over each run's best five checkpoints. The practical frontier is dominated by larger judges and Answer-Only rewarders, with the 8B Thinking judge remaining competitive.}
\label{tab:practical-recipe}
\end{table}

\subsection{Effect of Policy Size}
\researchquestion{Does scaling the policy improve quality, and does the effect differ between Answer-Only and Thinking supervision?}
\begin{figure}[t]
    \centering
    \subfloat[Answer-Only GenRMs]{\includegraphics[width=0.47\textwidth]{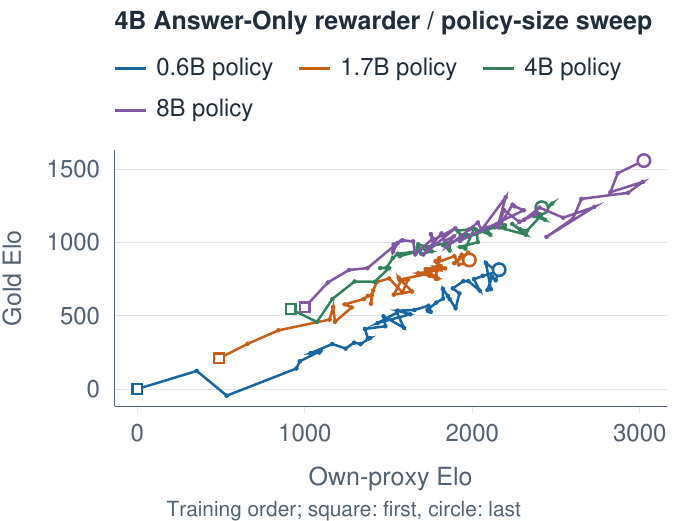}}
    \hfill
    \subfloat[Thinking GenRMs]{\includegraphics[width=0.47\textwidth]{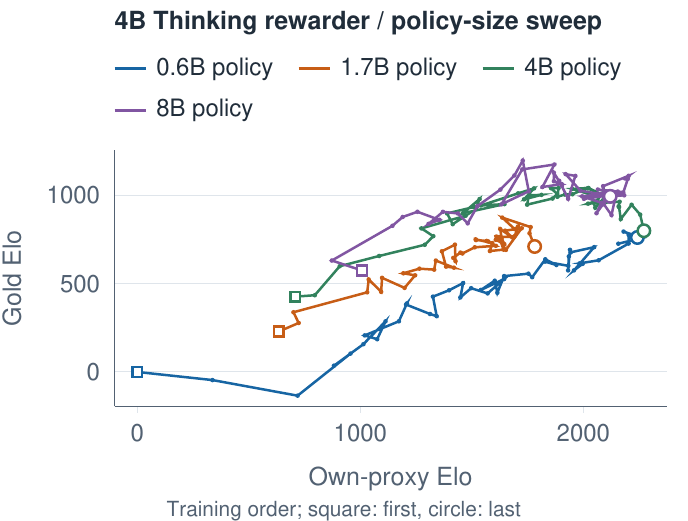}}
    \caption{Policy-size sweeps with a fixed 4B rewarder, in the retained family-specific arenas. Dots follow training chronology; squares/circles mark first/last observations. Gold and own-proxy Elo are separately fitted, and heights across the two panels are not one common arena. Larger policies improve under Answer-Only supervision in this sweep; larger Thinking-supervised policies reach their Gold peak earlier.}
    \label{fig:policy-size}
\end{figure}
We examine the effect of scaling policy size while fixing the GenRM.
Figure~\ref{fig:policy-size} shows results for policies ranging from 0.6B to 8B trained with both Answer-Only and Thinking GenRMs.
For Answer-Only supervision, larger policies achieve higher Gold Elo across the studied scales. This measures increased quality under the learned reference, not independent evidence that larger policies generalize better to human preferences.

Under Thinking supervision, the largest policy achieves the highest peak in the policy-size arena but declines earlier. The 0.6B and 1.7B policies peak near the end of the recorded window, the 4B policy around step 320, and the 8B policy around step 130. Larger policies may exploit a rewarder's weaknesses sooner, but these trajectories do not identify a mechanism or prove that additional policy capacity is intrinsically harmful. They show why judge capacity and policy capacity must be examined together, including training duration rather than only peak ratings.

\subsection{Effect of the \texorpdfstring{$\beta$}{beta} Coefficient in Online DPO}
\begin{figure}[htbp]
    \centering
    \includegraphics[width=\textwidth]{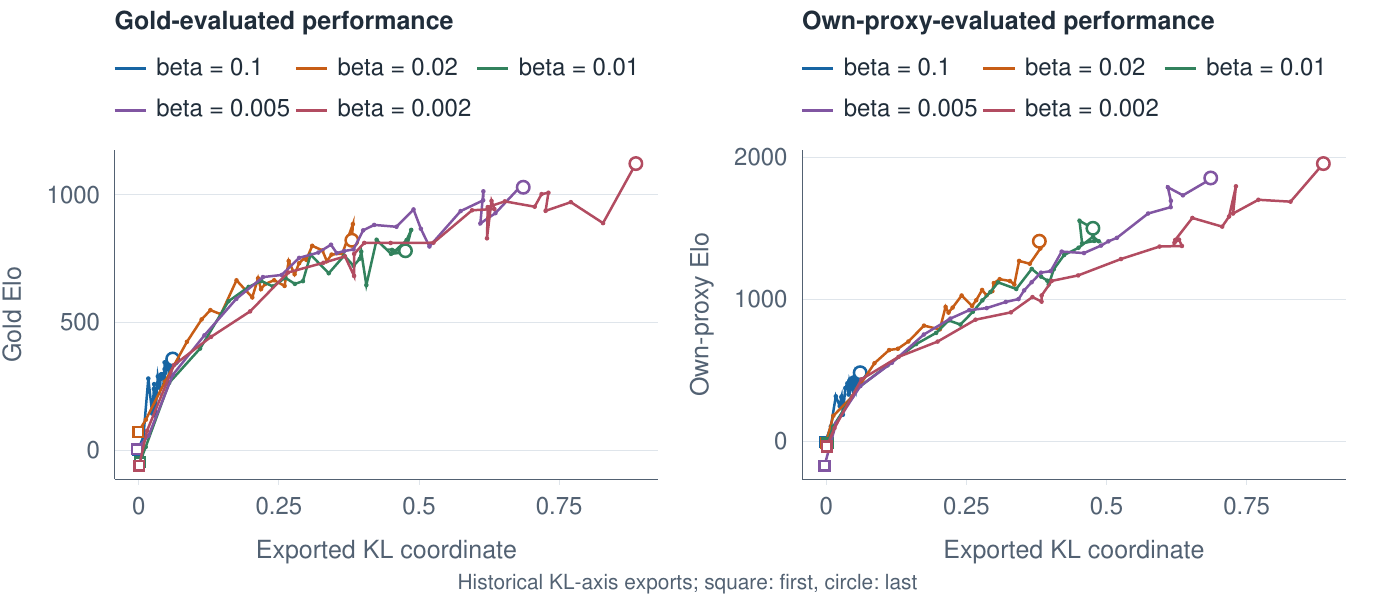}
    \caption{Gold and own-proxy ratings against historically smoothed KL export coordinates for the $\beta$ sweep. Lines connect exported observations in training order. The final step-260 rating has no retained KL coordinate and is omitted for each condition; no coordinate is synthesized. Negative exported values remain visible, not interpreted as raw nonnegative-estimator measurements.}
    \label{fig:beta-kl}
\end{figure}

\researchquestion{How does DPO regularization affect Gold ratings, and what do the retained KL exports show?}
\begin{figure}[t]
    \centering
    \includegraphics[width=0.68\textwidth]{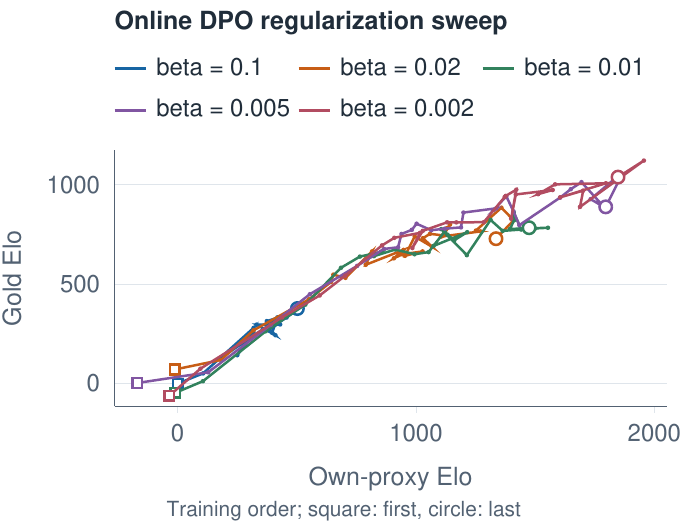}
    \caption{Online-DPO $\beta$ sweep under an Answer-Only rewarder. Points are retained snapshots connected in training order, not fitted smoothers; squares/circles mark first/last observations. Proxy and Gold Elo are separate fits. Smaller tested $\beta$ values reach higher Gold ratings, without strict monotonicity. The original account specifies 40{,}000 training prompts for this sweep.}
    \label{fig:beta}
\end{figure}
We analyze the effect of the $\beta$ coefficient in Online DPO.
Figure~\ref{fig:beta} shows proxy versus Gold Elo for $\beta \in \{0.1, 0.02, 0.01, 0.005, 0.002\}$. The smallest tested value, $\beta=0.002$, gives the highest Gold Elo. The ordering is not strictly monotone: $0.02$ slightly exceeds $0.01$ in the complete stored trajectory.
Lower $\beta$ does not simply increase proxy ratings at Gold's expense: Gold performance also improves over the observed range.
Figure~\ref{fig:beta-kl} shows Gold and own-proxy ratings against retained KL export coordinates. The original beta-plot path applies Savitzky--Golay smoothing before exporting these coordinates; they are not raw per-snapshot KL measurements. Smaller tested $\beta$ values reach higher Gold ratings over this window, but the plot does not establish a precise KL-constrained frontier or a safe divergence threshold.
The result favors smaller $\beta$ within the tested range, without identifying an optimum outside it. In the exported-KL subset, peak Gold Elo rises from 357 at $\beta=0.1$ to 1122 at $\beta=0.002$, paired with export coordinates of 0.061 and 0.887. The complete stored $\beta=0.1$ rating trajectory reaches about 378 Elo at step 260, beyond that subset. Negative export coordinates can arise from preprocessing; they are not evidence of negative true KL or attributed to finite-sample negativity of the released nonnegative estimator.

\section{Limitations and Discussion}
\label{sec:limitations}

\textbf{Reference supervision and validation reuse.} Gold is a 32B model fine-tuned on implicit, vote-derived preferences over human-written Reddit stories. Proxy labels then come from Gold, retaining nonzero sums of order-normalized verdicts; this can retain a pair with one vote parsed as invalid and does not eliminate reference-model bias. Gold validation, proxy validation and policy validation reuse the same 1{,}000 source pair records and their prompts, with different labels or generated responses. These measurements therefore do not constitute independent source-set replications. The downstream result is optimization toward a specified learned reference, not direct human adjudication of generated stories.

\textbf{Evaluator dependence and response quality.} Gold uses an Answer-Only Qwen3 model, and proxies share its model family and derived labels. Baseline comparisons do not rule out evaluator-specific preferences. A retained follow-up audit used a different learned judge, \auditjudge{}, on 40 randomly selected validation prompts from each of two 4B-policy settings, comparing each mode's peak-Gold outputs with randomized presentation order. Answer-Only won 42 of 80 comparisons (52.5\%; 95\% Wilson interval 41.7--63.1\%): 17/40 with 4B rewarders and 25/40 with 14B rewarders. This limited audit is inconclusive overall and does not independently establish the Gold-ranked advantage. It uses another Qwen3 Answer-Only judge, a different rubric and Gold-selected checkpoints, not human evaluation or a cross-family replication. Appendix~\ref{app:independent-judge} gives the procedure. \textit{Takeaway.} The main result is conditional on the specified learned Gold reference; neither format neutrality nor universal writing-quality superiority is established.

\textbf{Distribution-dependent influence of judge fine-tuning.} One unresolved hypothesis is that, on shifted policy outputs, fine-tuned judges increasingly share the preferences of their original models. If the learned Answer-Only Gold evaluator and Answer-Only proxies share such priors, part of the measured advantage could reflect reference alignment rather than human preference. The retained Elo trajectories do not test this mechanism. A discriminating check would compare Gold with its own unfine-tuned 32B parent on identical human-story and early/late policy response pairs, controlling response order and output validity. Agreement alone would still not establish human quality or its decline. We report this as a hypothesis, not a result or an explanation established by this study.

\textbf{Domain scope.} The trained-rewarder sweeps use creative writing; we do not systematically evaluate safety, helpfulness or verifiable tasks such as math and code. An ancillary OpenOrca experiment optimized policies using off-the-shelf rewarders and a different reference evaluator (Appendix~\ref{app:openorca}). It does not replicate the trained creative-writing rewarders or establish transfer of the central result.

\textbf{Relation to prior creative-writing studies.} Optimizing preference win rate can reduce stylistic diversity \citep{chung2025diverse}; the retained generations show style convergence under both judge modes. \citet{fein2025litbench} report that chain-of-thought can degrade verification accuracy for creative writing, and that trained BT and generative verifiers outperform zero-shot judges on their benchmark. Our 4B trace-distillation ablation gives a smaller static-accuracy gain than GRPO, and trained Thinking judges are the stronger static evaluators in our sweep. Their downstream reversal therefore cannot be inferred from a generic claim that thinking is unsuitable for creative-writing evaluation. It concerns the complete training and measurement loop.

\textbf{Family and algorithm scope.} All models use Qwen3 backbones; family-specific priors could affect the observed rankings and size slopes. Cross-family work also shows that behavioral priors can change RL outcomes \citep{gandhi2025stars}. We do not compare PPO-style policy optimization, alternative Thinking-judge training recipes or Best-of-N inference-time selection. The study concerns generative reward models as online-DPO training signals, not a compute-matched ranking of all ways to use a judge.

\textbf{Elo anchoring and schedule.} Ratings depend on evaluator, contestants, anchor and match schedule \citep{chatbot-arena-2024}. Proxy and Gold ratings are fitted separately; the diagonal of a proxy--Gold plot is not a verdict-agreement statistic. We report global and size-stratified arenas. Bootstrap intervals for Answer-Only-minus-Thinking gaps independently resample stored winner/loser rows and refit a lightly regularized Bradley--Terry model. They are conditional on these stored comparisons, not robust to arbitrary dependence between rows sharing prompts or responses, and do not capture fresh generation, retraining, seed, anchor or tournament-design uncertainty. \textit{Takeaway.} Interpret within-arena differences alongside these dependence and selection limits, not as universal confidence statements.

\textbf{Post-hoc fits and KL provenance.} Appendix~\ref{app:scaling-law-fits} fits model-size associations to the stored grid, not seed-averaged laws or out-of-sample predictions. Retained KL coordinates come from historical analysis exports, with smoothing established for the beta plot and incomplete preprocessing provenance for other panels. Matching them to exported Gold ratings does not independently validate the original logger-to-step mapping. We preserve these coordinates for inspection but do not use them to identify a quantitative KL-constrained frontier.

\section{Related Work}

Reward overoptimization is studied in RLHF \citep{ouyang2022training,gao2022scaling} and direct methods such as DPO \citep{rafailov2023dpo,rafailov2024rewardoveropt}. Prior evaluations explicitly distinguish static accuracy from downstream policy performance \citep{wen2025rethinking,frick2024evaluate,kim2025rethinking}; the generic distinction is not new to this work. GenRMs replace scalar-head prediction with text generation, allowing verdicts and rationales \citep{mahan2024generative,zhang2024genverifier}. Subsequent approaches include GRPO and positional consistency in J1 \citep{whitehouse2025j1}, self-principled critique and aggregation in DeepSeek-GRM \citep{liu2025inferencegrm}, and test-time aggregation \citep{wang2025helpsteer3,shi2025heimdall}. Related supervision uses generated critiques \citep{yu2025criticrm}, criteria trees \citep{liang2025syncpl} and reasoning plans \citep{saha2025evalplanner}; judge reliability and static benchmarks remain active topics \citep{ye2024llmasjudge,lambert2024rewardbench}.

Recent work also examines rationale-consistency supervision for GenRMs \citep{wang2026rationale} and rank-derived scalar rewards for GRPO \citep{wang2026rrc}. Our study instead uses direct pairwise preferences in online DPO to measure generative-rewarder capacity and practical Answer-Only SFT versus Thinking GRPO pipelines, with evaluator-specific Elo arenas \citep{chatbot-arena-2024}. It does not test those interventions, isolate a causal effect of reasoning format, or establish their mechanisms as explanations of our results.

\section{Conclusion}

We study generative reward models as online training signals. Under our learned Gold reference, Thinking GenRMs outperform Answer-Only variants on static evaluation, yet this advantage usually reverses during policy optimization. The result concerns the compared SFT/GRPO pipelines in this domain, not an isolated causal effect of reasoning or a universal ranking of writing quality.

Three findings offer practical guidance within these experiments: (1) reward model scale is a major driver of Gold-evaluated policy outcomes, with gains continuing even when the GenRM exceeds the policy in size; (2) intermediate GenRM checkpoints can outperform fully-trained models as rewarders; and (3) in-distribution accuracy does not by itself predict training effectiveness. Selecting GenRMs only by static benchmarks can therefore select checkpoints that are suboptimal for downstream optimization under the learned reference.

\bibliography{references}
\bibliographystyle{plainnat}

\clearpage
\appendix
\section{Empirical Trajectories and Capacity Sweeps}\label{app:empirical}

\subsection{Trained and Off-the-Shelf Rewarders over Training}

\begin{figure}[htbp]
  \centering
  \includegraphics[width=0.86\textwidth]{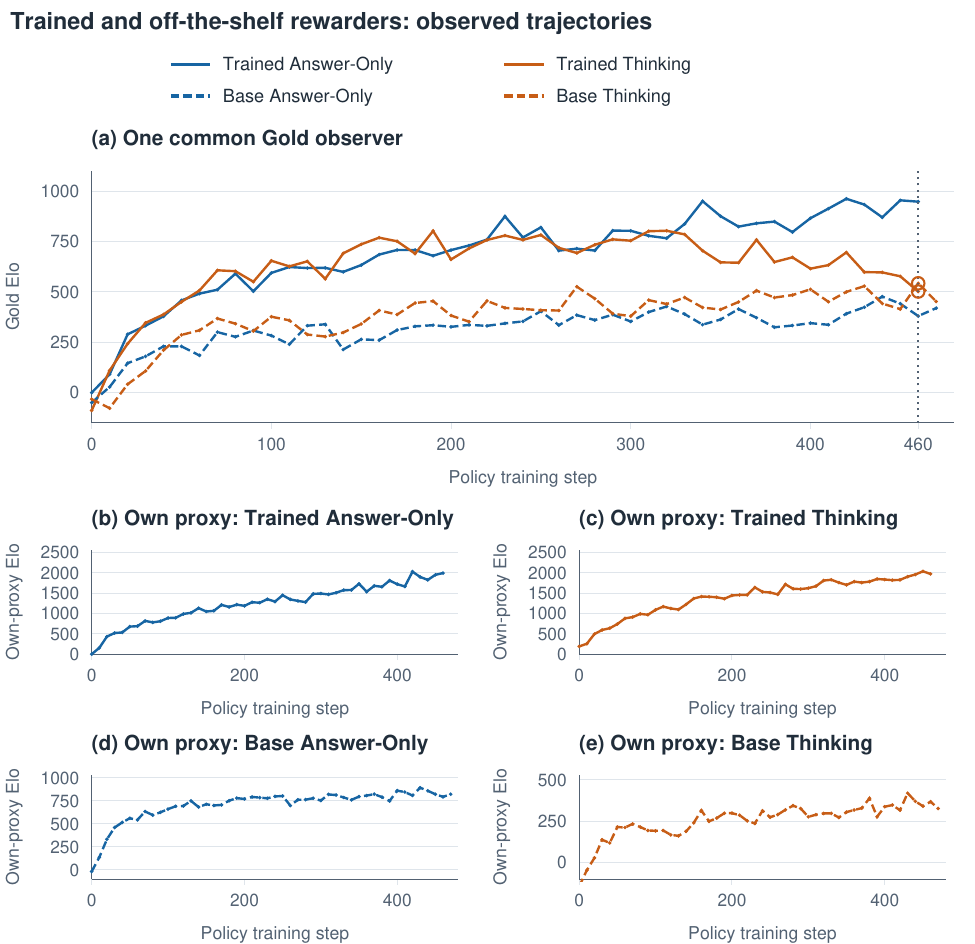}
  \caption{\textbf{Trained versus off-the-shelf rewarders through the retained
  training horizon.} Blue denotes Answer-Only and orange Thinking; solid lines
  denote trained rewarders and dashed lines off-the-shelf rewarders. The upper
  panel uses one Gold arena; outlined points mark step~460. Each lower panel
  uses its own rewarder's separately fitted ratings, so heights across those
  panels are not comparable. Policy and judge are 4B. Source:
  \texttt{trained\_vs\_baseline/*.csv}; no smoothing or missing-step interpolation.}
  \label{fig:restored-baseline-trajectories}
\end{figure}

Figure~\ref{fig:restored-baseline-trajectories} separates the common Gold
evaluation from the four rewarders' own-proxy ratings.

\begin{samepage}
Trained rewarders achieve higher Gold peaks in both formats. For Thinking,
the trained peak is~802.8 versus~542.3, but at shared step~460 the baseline
leads~542.3 to~501.5. At~450 the ordering is opposite (414.3 versus~577.1).
Only the baseline continues to~470. This late crossover is not a significance
result or uniform baseline superiority. From~320 to~460, the trained Thinking
policy's own-proxy rating rises~1808.3 to~1971.5 while Gold falls~802.8 to~501.5:
proxy--reference divergence, not an identified mechanism or human-quality decline.
\end{samepage}

\clearpage
\subsection{All Eight Matched Conditions in One Gold Arena}

\begin{figure}[H]
  \centering
  \includegraphics[width=\textwidth]{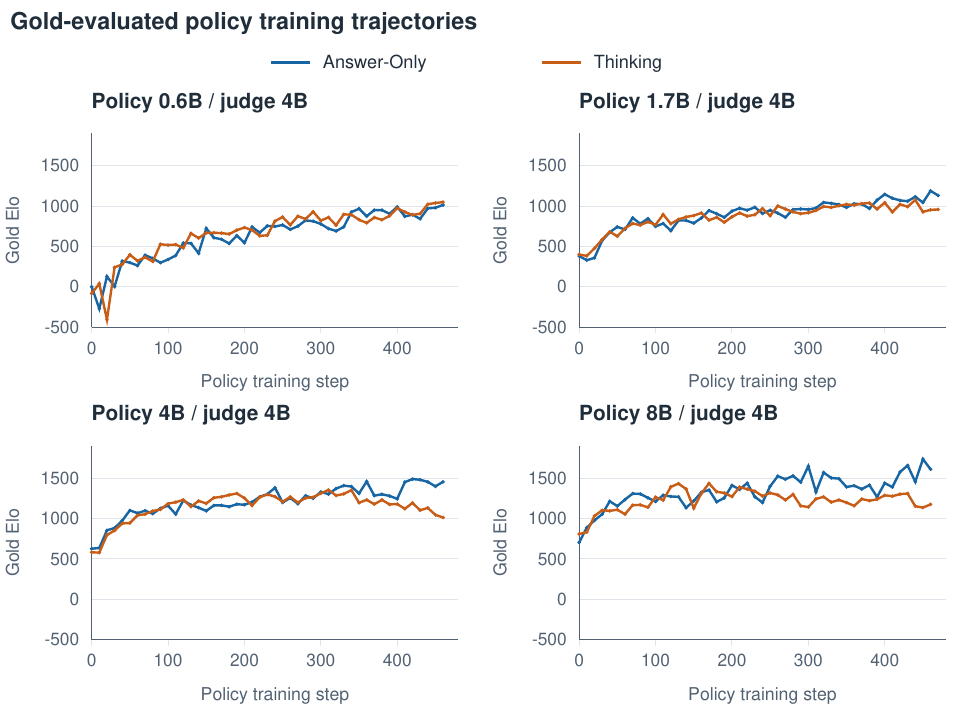}
  \caption{\textbf{Policy-size sweep with a fixed 4B rewarder.} All four panels
  use the common Gold arena, common vertical limits and observed policy-training
  steps. Blue: Answer-Only SFT rewarders; orange: Thinking GRPO rewarders.
  Each line represents one retained policy trajectory, not a mean across seeds.
  Source: \texttt{all\_params/gold.csv}.}
  \label{fig:restored-all-trajectories-a}
\end{figure}

\begin{figure}[H]
  \centering
  \includegraphics[width=\textwidth]{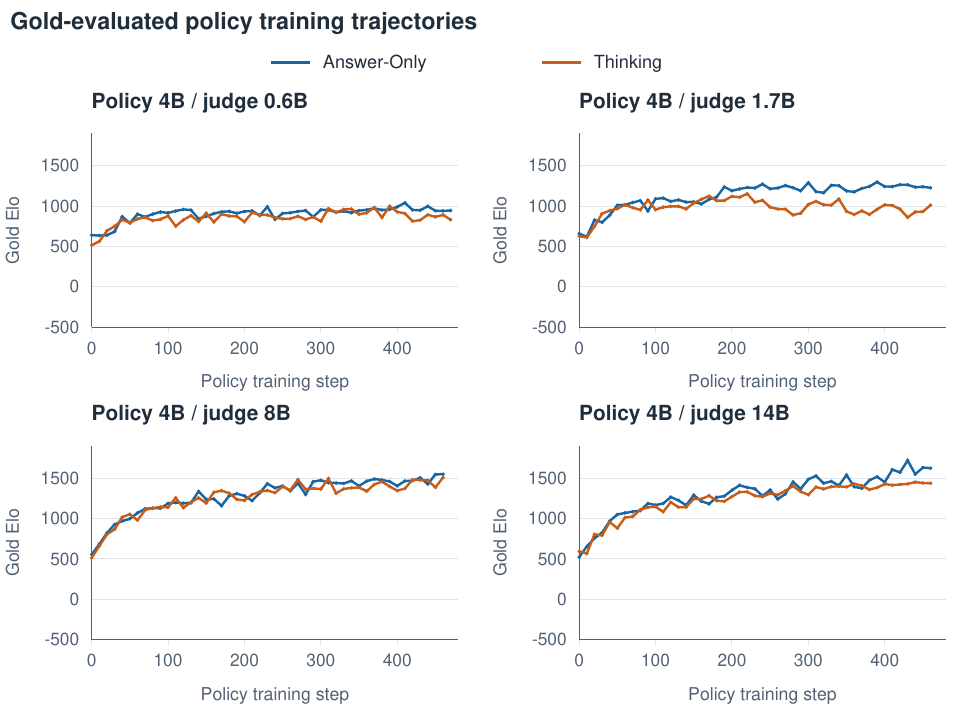}
  \caption{\textbf{Additional rewarder-size conditions with a fixed 4B policy.}
  Together with the 4B/4B panel of
  Figure~\ref{fig:restored-all-trajectories-a}, these are the five rewarder sizes
  studied. Colors, arena and axis limits match the preceding figure. The
  retained curves show measured behavior, not a claim of monotonic improvement
  at every step or across every rewarder size.}
  \label{fig:restored-all-trajectories-b}
\end{figure}

Figures~\ref{fig:restored-all-trajectories-a}
and~\ref{fig:restored-all-trajectories-b} show the complete stored Gold-rated
trajectories for all eight matched conditions. All panels use
\texttt{all\_params/gold.csv}, sharing one arena and vertical scale. They expose
early gains and late reversals hidden by best-checkpoint summaries; a falling
tail does not itself measure verdict agreement or Gold-to-base reversion.

\subsection{The Same Gold Observations against Exported KL Coordinates}

\begin{figure}[H]
  \centering
  \includegraphics[width=\textwidth]{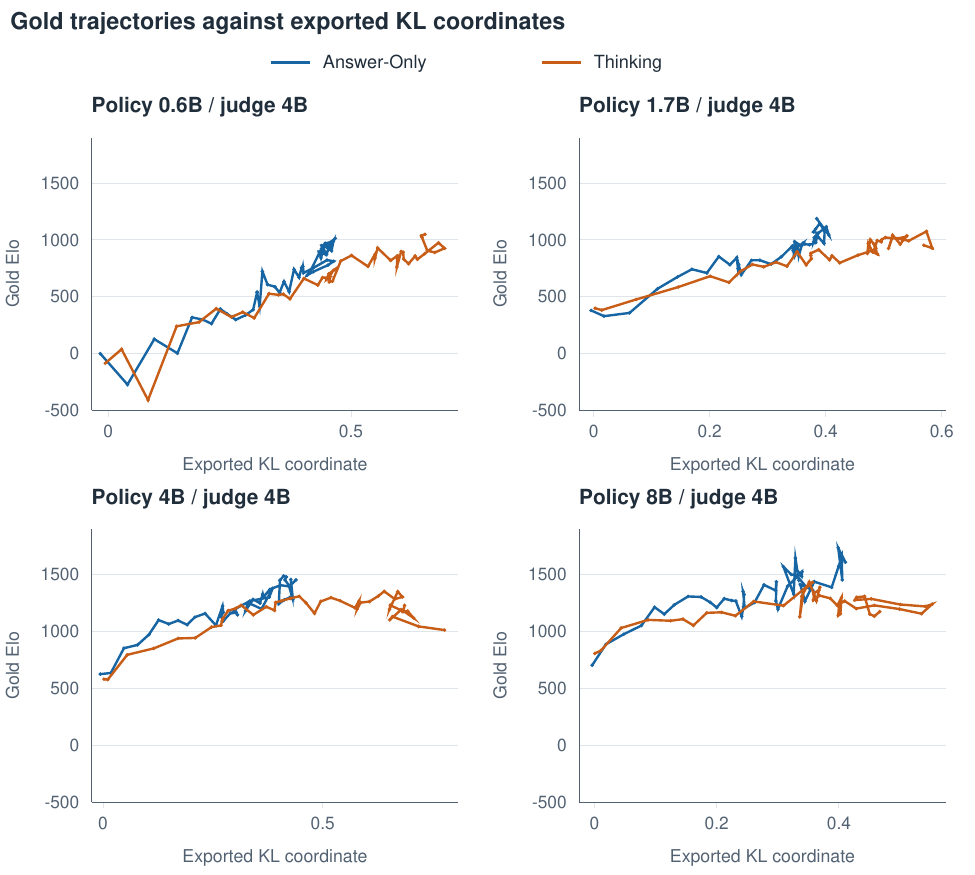}
  \caption{\textbf{Policy-size conditions against retained KL export coordinates.}
  Gold coordinates are the same common-arena observations as
  Figure~\ref{fig:restored-all-trajectories-a}; only the horizontal coordinate
  changes. Lines follow the recovered checkpoint order. Each cell has 47
  exported coordinates per mode; the two extra continuation snapshots in the
  1.7B-policy cell lack a retained KL coordinate and are excluded. Horizontal
  limits differ across cells; vertical limits and the Gold arena are shared.
  Original KL preprocessing is not fully reconstructed.}
  \label{fig:restored-kl-trajectories-a}
\end{figure}

\begin{figure}[H]
  \centering
  \includegraphics[width=\textwidth]{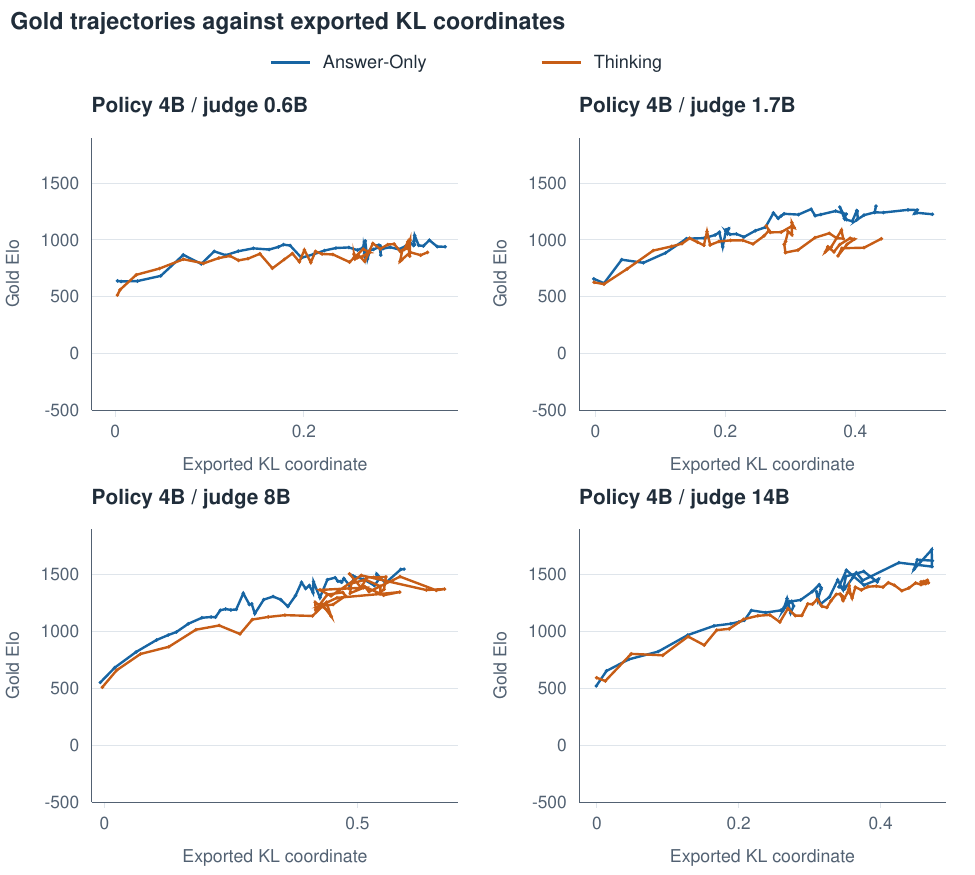}
  \caption{\textbf{Rewarder-size conditions against retained KL export coordinates.}
  Gold coordinates match Figure~\ref{fig:restored-all-trajectories-b}. Only
  uniquely matched exported observations are shown. The two extra
  continuation snapshots in the 0.6B-rewarder cell are absent from these KL
  exports and are excluded. Negative exported values remain visible without
  being interpreted as raw KL measurements. This is not a fitted KL scaling law.}
  \label{fig:restored-kl-trajectories-b}
\end{figure}

Figures~\ref{fig:restored-kl-trajectories-a}
and~\ref{fig:restored-kl-trajectories-b} provide a complementary view using the
retained policy-KL analysis exports rather than training step. The exports do
not contain explicit checkpoint IDs. For each exported Gold row, we require a
\emph{unique} match on its exact trajectory prefix and Gold rating (absolute
tolerance~$10^{-7}$) in the common-arena file. This recovers 752 exported rows.
Four additional global-arena snapshots have no matching KL coordinate and are
omitted, not assigned an inferred value. The machine-readable join ledger records
every source row and recovered checkpoint.

These coordinates are preserved exports, not newly recovered raw per-step KL.
The original analysis code smooths KL before export for the beta comparison;
the exact preprocessing path for these matched panels is not fully reconstructed.
The released trainer's k3 estimator is nonnegative per token in exact arithmetic,
so negative exported coordinates are not attributed to ordinary finite-sample
negativity of that estimator. They are retained rather than clipped. Our unique
rating match identifies the exported observation, not a new validation of the
original positional logger-to-step mapping. Lines follow the recovered rating
checkpoint order even when a KL coordinate moves backward. These panels are
ancillary descriptive views, not measurements of distance from Gold's training
distribution or a validated quantitative KL budget.

\subsection{Rewarder and Policy Capacity Sweeps}

\begin{figure}[H]
  \centering
  \includegraphics[width=\textwidth]{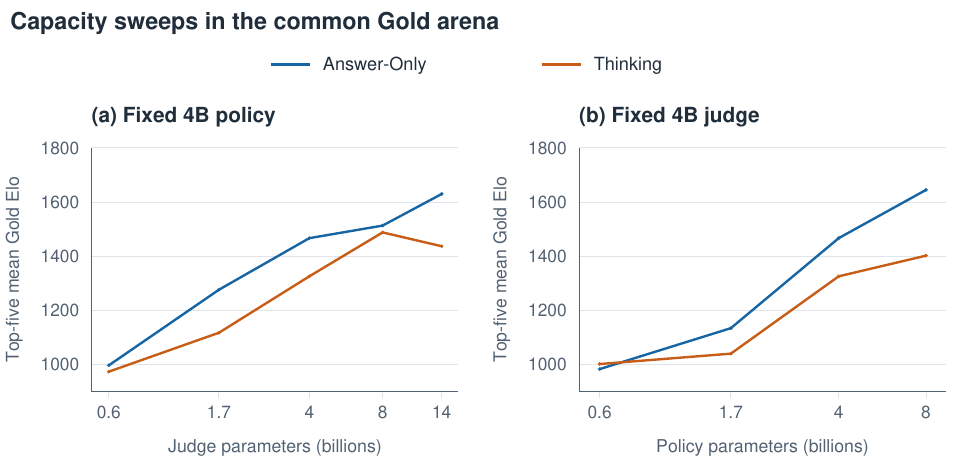}
  \caption{\textbf{Observed capacity sweeps in the common Gold arena.}
  Left: rewarder size varies while the policy is 4B. Right: policy size varies
  while the rewarder is 4B. Each point is the average of the five highest
  recorded Gold-rated snapshots of that trajectory. Parameter-size axes are
  logarithmic and labeled with actual sizes. Blue: Answer-Only; orange:
  Thinking. These are descriptive observed summaries, not compute-normalized
  optima, predictive scaling fits, or independent-seed confidence intervals.}
  \label{fig:restored-capacity}
\end{figure}

Figure~\ref{fig:restored-capacity} preserves both sweep directions: rewarder
size at fixed policy size, and policy size at fixed rewarder size. The mean
of the five highest Gold-rated snapshots is still selected under Gold, not an
independently validated deployment rule. Only observed cells are shown.

Even parameter-count-matched swaps need not favor the same allocation in both
formats. At a 12B summed parameter count, the 8B-policy/4B-rewarder condition has
Answer-Only top-five Gold Elo~1645.0, versus~1513.1 for the 4B-policy/8B-rewarder
condition. Thinking reverses that ordering (1402.2 versus~1488.1). A parameter
sum is not an inference- or training-compute budget, especially when one rewarder
generates reasoning traces; these observed swaps do not identify a universal
compute-optimal allocation.

\paragraph{Intermediate judge training.}
Figure~\ref{fig:intermediate-judges} retains the intermediate-Thinking-checkpoint
ablation. The original account specifies judge steps 0, 50, 100, 150, 200,
250 and 300 and a 40,000-policy-prompt budget. We preserve that image rather
than infer static accuracies or checkpoint identities from pixels; its source
account does not substitute for an exact per-judge run manifest.

\paragraph{Reconstruction and provenance.}
The \href{https://github.com/RaychevAngel/generative-reward-models/tree/main/analysis}{released analysis scripts}
and their provenance ledgers record source hashes, row counts,
exported-coordinate joins, omissions and condition summaries. These six
appendix figures introduce no model runs, new labels or significance tests.

\section{Evaluation and Analysis Details}

\subsection{Elo Estimation and Conditional Uncertainty}
\label{app:elo}

For each evaluator and arena, we fit Bradley--Terry ratings from its stored pairwise winner/loser records. With checkpoint ratings $R_i$, the fitted win probability is
\[
p_R(i\succ j)=\sigma\!\left(\frac{R_i-R_j}{s}\right),
\qquad s=\frac{400}{\log 10}.
\]
The anchored log-likelihood sums $\log p_R(i\succ j)$ over observed wins, with one reference rating fixed at zero. Each evaluator and contestant set has a separate fit. Additive anchors do not affect within-arena differences, but ratings from different arenas cannot simply be subtracted. In particular, the diagonal of a proxy--Gold plot marks equal numerical ratings, not identical preferences or calibrated agreement.

For the reported uncertainty analysis, we independently resample stored winner/loser rows and refit the ratings 200 times. The refits add an L2 penalty $\frac{0.01}{2}\|\theta_{-s_0}\|^2$ in logit coordinates $\theta=R/s$, with reference $s_0$ fixed. They are therefore closely related to, but not byte-identical with, the original exported ratings. For configuration $c$ we report
\[
\mathrm{Peak}(c)=\max_t R_{c,t},
\qquad
\mathrm{Top5}(c)=\frac{1}{5}\sum_{t\in\mathcal T_5(c)}R_{c,t},
\]
where $\mathcal T_5(c)$ contains that run's five highest-rated snapshots. Each resample repeats this selection before computing Answer-Only-minus-Thinking gaps; the reported 95\% intervals are bootstrap percentiles.

These intervals describe sensitivity to the stored match table under an independent-row approximation. Comparisons sharing prompts or responses, or edges broken from one partial ranking, can be dependent. The retained winner/loser table does not preserve the cluster identifiers needed for a prompt-aware bootstrap. The intervals do not include fresh generations, independent training seeds, judge changes or tournament redesign. Peak and top-five selection use the evaluation data and should not be read as prospective checkpoint-selection guarantees.

\subsection{Descriptive Scaling Fits}
\label{app:scaling-law-fits}

We summarize the stored sweep with post-hoc fits, not a theory identifying neural optimization with a cardinal-utility model. For policy size $P_c$, rewarder size $R_c$ (billions of parameters) and Thinking indicator $M_c$, ordinary least squares fits
\[
Y_c=b_0+b_P\log P_c+b_R\log R_c+b_M M_c
       +b_{MP}M_c\log P_c+b_{MR}M_c\log R_c+\epsilon_c,
\]
with $Y_c$ the peak or top-five Gold Elo. There are 16 matched mode/size configurations and six fitted coefficients. The interaction terms permit different size slopes for the two pipelines; they do not isolate response mode from its training objective.

\begin{table}[htbp]
\centering\small
\setlength{\tabcolsep}{4pt}
\begin{tabular}{lrrrrrr}
\toprule
Metric & $R^2$ & RMSE & Ans.\ policy & Ans.\ judge & Think policy & Think judge \\
\midrule
Peak Elo & 0.954 & 51.8 & 186.0 & 144.7 & 113.3 & 113.3 \\
Top-5 Elo & 0.952 & 51.2 & 175.8 & 135.7 & 119.0 & 115.6 \\
\bottomrule
\end{tabular}
\caption{In-sample log-size fits to the 16 matched global configurations. Slope columns are fitted Gold-Elo differences per parameter doubling along each axis, not observed new doublings, compute-matched gains or out-of-sample predictions. RMSE is in Elo units.}
\label{tab:scaling-law-fits}
\end{table}

Both policy and judge scale have substantial fitted associations with Gold Elo in this grid. The high in-sample $R^2$ does not establish a universal scaling law: the grid is sparse, checkpoints were selected by the same learned evaluator and no independent-seed uncertainty is included.

Inputs, coefficients and the model-size fitting script are included in the public research artifacts. Earlier exploratory KL-curve fits remain in the research archive rather than serving as quantitative scaling evidence here: the exported horizontal coordinates include historical preprocessing whose complete provenance is not reconstructed.

\subsection{Why Static Accuracy Is Not an Online-Update Guarantee}
\label{app:dpo-noise}

A short update-level identity helps interpret the empirical distinction. Let $\Delta_\theta$ denote the log-probability margin, relative to the reference policy, for a fixed Gold-oriented response pair. The Gold-label loss is $\ell_+=-\log\sigma(\beta\Delta_\theta)$; a flipped label gives $\ell_-=-\log\sigma(-\beta\Delta_\theta)$. Since $\sigma(u)+\sigma(-u)=1$,
\[
\nabla_\theta\ell_- - \nabla_\theta\ell_+
=\beta\nabla_\theta\Delta_\theta.
\]
Thus two rewarders with the same average error can induce different updates if their mistakes occur on pairs with different gradient directions or magnitudes. The relevant pairs are sampled from the evolving policy, not the fixed static validation distribution.

Filtering adds a second difference. The actual trainer sums order-normalized verdicts and retains nonzero-sum pairs; votes parsed as invalid contribute zero. Different proxies can therefore change both the labels of retained pairs and the distribution of pairs that survive. High static accuracy alone controls neither effect. This is a possible explanation of the evaluator--rewarder gap, not a measured decomposition: the retained artifacts do not supply the per-pair margins and gradient features needed to estimate it, nor establish which judge's errors better track human preferences.

\subsection{Follow-Up Audit with a Different Learned Judge}
\label{app:independent-judge}

We report an existing follow-up diagnostic separately from the main Gold arena. For 4B policies trained with 4B and 14B rewarders, we selected each mode's peak-Gold snapshot and sampled 40 validation prompts per setting with seed 17. Response presentation order was randomized. The judge, \auditjudge{}, received the prompt and two responses without their training-mode labels. It judged prompt-following, coherence, specificity, originality and writing quality, with an instruction not to favor length. It produced one Answer-Only verdict per pair at temperature zero with a 24-token limit; no swapped-order repeat was performed. All 80 verdicts were parseable.

\begin{table}[ht]
\centering\small
\begin{tabular}{lrrl}
\toprule
Rewarder size (4B policy) & Answer-Only wins & Comparisons & 95\% Wilson interval \\
\midrule
4B & 17 & 40 & 28.5--57.8\% \\
14B & 25 & 40 & 47.0--75.8\% \\
Combined & 42 & 80 & 41.7--63.1\% \\
\bottomrule
\end{tabular}
\caption{A different learned judge does not establish an overall preference advantage in this limited diagnostic. These descriptive binomial intervals are not prompt-cluster or retraining intervals; the 80 comparisons contain 76 unique prompts, with four prompts shared across settings.}
\end{table}

The diagnostic is inconclusive overall. Its two settings, selected checkpoints, changed rubric, shared Qwen3 family and single verdict per pair limit interpretation. It neither reproduces a human evaluation nor supplies a thinking-style Gold control; it makes the evaluator dependence of the main result explicit.

\subsection{Ancillary Policy Optimization on OpenOrca}
\label{app:openorca}

We retain an ancillary OpenOrca policy-optimization comparison. The recovered launch and plotting code identifies a 1.7B SFT-initialized policy followed by online DPO with off-the-shelf Qwen3 rewarders across several sizes and response modes. Its separate reference evaluator is a 32B Thinking model using a general response-quality rubric, not the learned Answer-Only Gold evaluator used for creative writing.

The retained plot reports per-step net preference rates against fixed reference responses: the number of wins minus losses, divided by the number of comparisons. Each point summarizes an evaluated policy state under its training rewarder and the separate reference evaluator. The exact numerical source-table revision for this image has not been reconstructed. We preserve it as ancillary historical evidence, not a new common-arena rating analysis, direct verdict-agreement measurement or replication with trained creative-writing GenRMs.

\begin{figure}[H]
    \centering
    \includegraphics[width=0.85\linewidth]{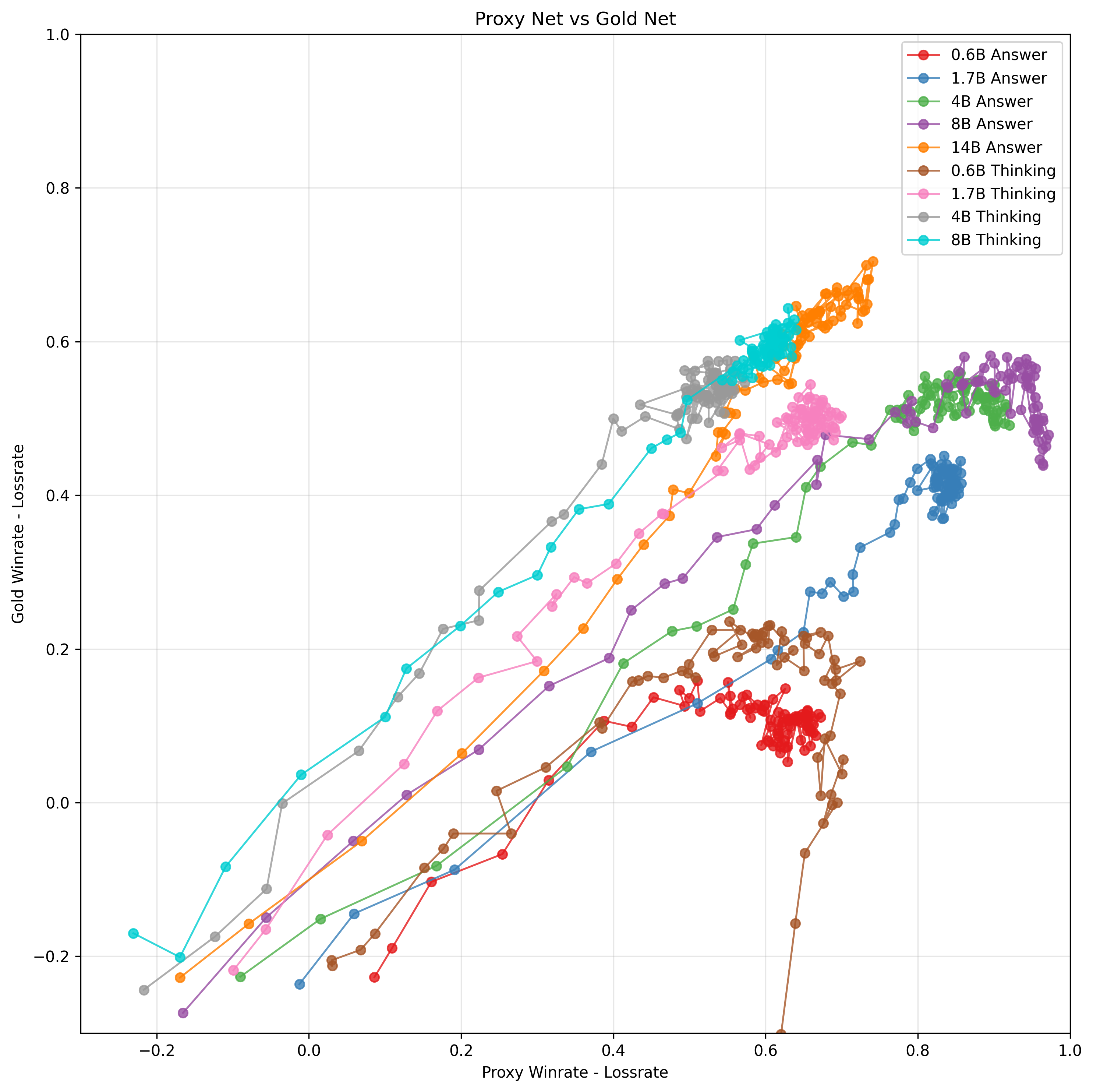}
    \caption{Retained OpenOrca policy-optimization trajectories with off-the-shelf rewarders. Axes show net preference rates, not Elo. Recovered source identifies online DPO from an SFT-initialized policy and a separate 32B Thinking reference evaluator. This is not a trained-GenRM cross-domain replication; exact numerical reconstruction of the retained image remains unavailable.}
    \label{fig:openorca}
\end{figure}

\clearpage
\section{Prompt Templates}
\label{sec:prompts}

We provide the retained creative-writing prompt templates for policy generation and GenRM pairwise evaluation. Models apply their own tokenizers and chat templates; Thinking and Answer-Only settings additionally control whether a reasoning trace is generated. The ancillary OpenOrca experiment uses a different task prompt and reference rubric.

\subsection{Policy Prompt}

The policy model receives the following prompt template for creative writing generation:

\begingroup\small
\begin{verbatim}
Creative Writing Prompt:
{QUESTION}
Goal: write an original short story that directly answers the prompt.
Requirements: Output only the story text. no titles, notes, commentary
or meta-text.
\end{verbatim}
\endgroup

\noindent The \texttt{QUESTION} placeholder is replaced with the specific creative writing instruction from the evaluation set.

\subsection{GenRM Pairwise Evaluation Prompt}

The generative reward model receives the following prompt template for pairwise comparison:

\begingroup\small
\begin{verbatim}
Two AI models were given the same instruction and each produced a reply.
Your task is to judge which reply better fulfills the instruction.

[BEGIN INSTRUCTIONS GIVEN TO BOTH MODELS]
{POLICY_PROMPT}
[END INSTRUCTIONS]

[BEGIN REPLY A]
{RESPONSE_A}
[END REPLY A]

[BEGIN REPLY B]
{RESPONSE_B}
[END REPLY B]

Output ONLY one of:
<answer>[[A]]</answer>
<answer>[[B]]</answer>
\end{verbatim}
\endgroup

\noindent The placeholders are filled with the full policy prompt and the two candidate responses. The model outputs a structured verdict indicating which response better satisfies the original instruction.

\end{document}